# A review on Neural Turing Machine

Soroor Malekmohammadi Faradounbeh, Faramarz Safi-Esfahani

Faculty of Computer Engineering, Najafabad Branch, Islamic Azad University, Najafabad, Iran.
Big Data Research Center, Najafabad Branch, Islamic Azad University, Najafabad, Iran.
soroormalekmohamadi@gmail.com, fsafi@iaun.ac.ir

*Abstract*— one of the major objectives of Artificial Intelligence is to design learning algorithms that are executed on a general purposes computational machines such as human brain. Neural Turing Machine (NTM) is a step towards realizing such a computational machine. The attempt is made here to run a systematic review on Neural Turing Machine. First, the mind-map and taxonomy of machine learning, neural networks, and Turing machine are introduced. Next, NTM is inspected in terms of concepts, structure, variety of versions, implemented tasks, comparisons, etc. Finally, the paper discusses on issues and ends up with several future works.

## 1  INTRODUCTION

The Artificial Intelligence (AI) seeks to construct real intelligent machines. The neural networks constitute a small portion of AI, thus the human brain should be named as a Biological Neural Network (BNN). Brain is a very complicated non-linear and parallel computer [1]. The almost relatively new neural learning technology in AI named 'Deep Learning' that consists of multi-layer neural networks learning techniques. Deep Learning provides a computerized system to observe the abstract pattern similar to that of human brain, while providing the means to resolve cognitive problems. As to the development of deep learning, recently there exist many effective methods for multi-layer neural network training such as Recurrent Neural Network (RNN)[2].

What is possible as a reality is not so simple in practice. Computation programs are constructed based on three fundamental mechanisms: 1) Initial operations (e.g. arithmetic), 2) Logical flow control, (e.g. branching) and 3) External memory, to allow reading and writing (Von Neumann, 1945). With respect to the success made in complicated data modeling, machine learning usually applies logical flow control by ignoring the external memory. Here, RNNS networks outperform other learning machine methods with a learning capability. Moreover, it is obvious that RNNS, are Turing-Complete[3] and provided that they are formatted in a correct manner, they would be able to simulate different methods. Any advance in RNNS capabilities can provide solutions for algorithmic tasks by applying a big memory. A factual sample of this is the Turing enrichment where it benefits from limited state machines through an unlimited memory strip [4, 5].

In human cognition, a pattern resembling to electronic operation is named Working Memory, the structure of which has remained as an ambiguous issue and neuron-physiology level. In fact, this memory is a capacity to reserve short-term information and apply information based on comprehensive rules [6-8]. In 2014, Neural Turing machine similar to a working memory system is introduced with the potential to access data based on determined regulations.

Attempts are made here to run a review on the newly introduced methods in the field of deep learning, like the neural Turing machine (NTM). In this paper, the concepts of machine learning, Turing machine, and neural network are reviewed followed by reviewing the newly introduced methods and finally their comparison.

## 2    LITERATURE REVIEW

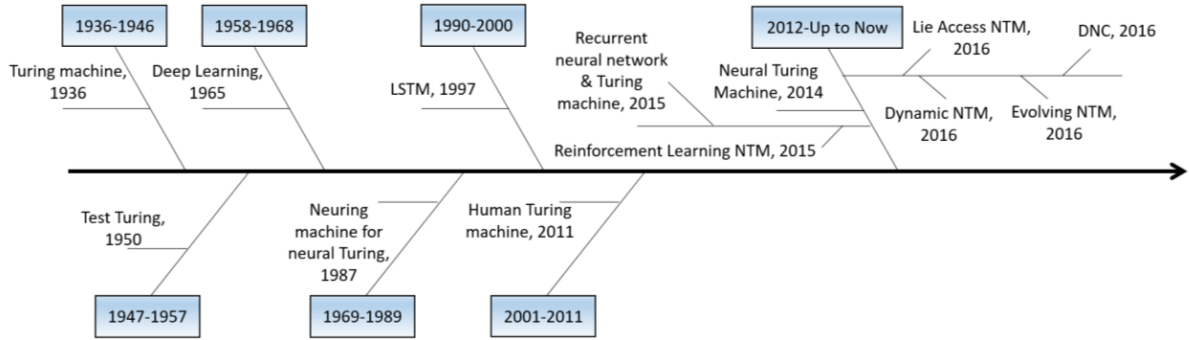

*Fig. 1. The chronology of NTM*

As observed in Fig.1, NTM is of a seventy years background, coinciding the construction of the first electronic computer [9]. Designing all purposes algorithms has always been and is the objective of researchers in this field [10]. In 1936, the first machine was invented for this purpose by Alan Turing and was named the Turing Machine. In its initial days, its commission affected the psychological and philosophical concepts to a degree that computer was considered as a simulation of brain function, thus a metaphor. This precipitation was rejected by the experts in neuron related science since there is no resemblance between the Turing machine architecture and that of human brain [11].

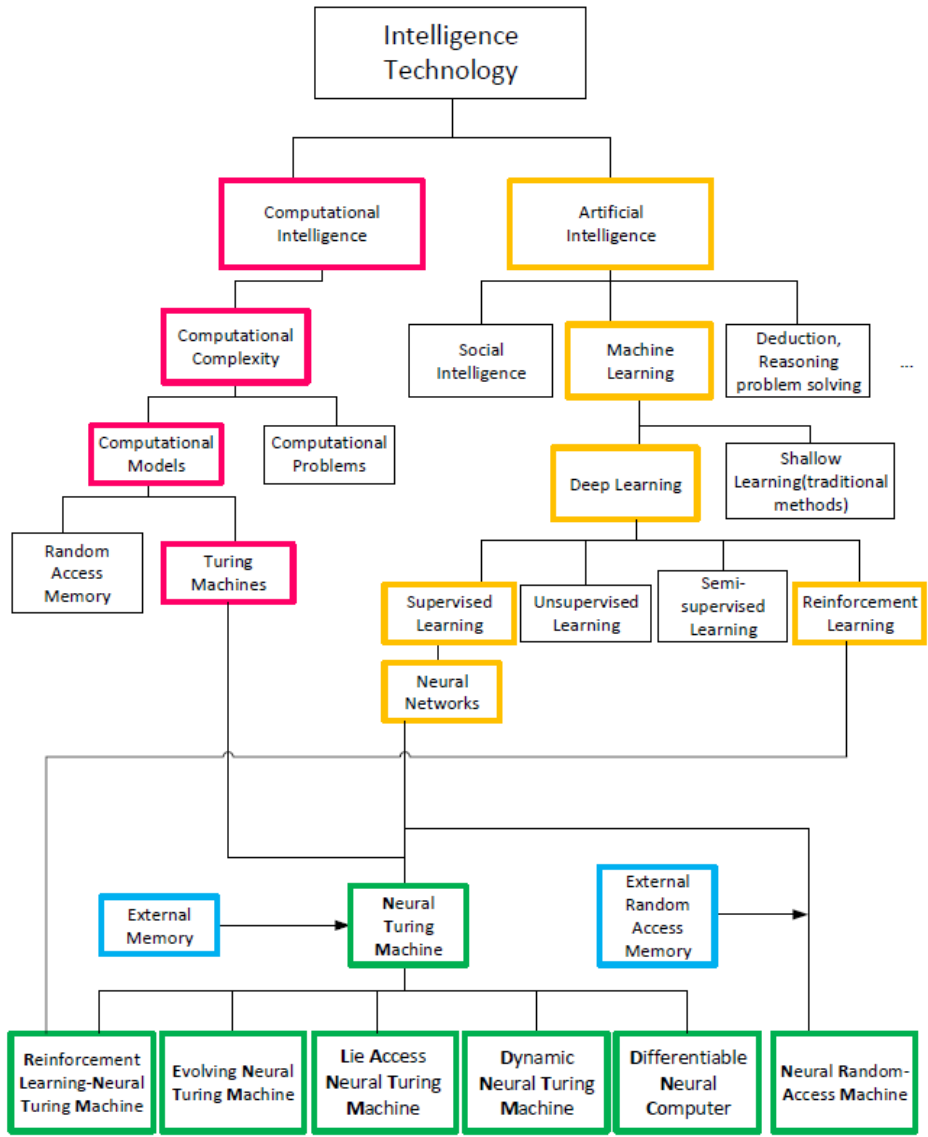

*Fig. 2. The mind map of this article*

## 2.1 Machine Learning, Deep, Reinforcement, and Q learning

It can be claimed that learning process is materialized when the subject becomes able to perform in another manner or probably much better based on the gained experiences. In fact machine learning is involved in studying computerized algorithms for learning how to deal with problems [12]. The purpose of machine learning, is not only to realize the main learning process unique to human and animal, but also to construct feasible systems in any engineering field. The objective of learning is finding manners of function in different states [13]. In general, the main objective of most studies regarding machine learning is to develop algorithms of scientific worth for general application [12-14]. As observed in Fig. 2, machine learning is a sub-set of AI divisible in two shallow and deep learning sections. Here, focus is on deep learning section.

The objective of Deep learning is defined as approaching one of the main objectives, that is, AI[1]. The Deep Neural Network, including the feedforward neural networks, have been and are successful in determining machine learning patterns [15]. Said otherwise, deep learning is a class of machine learning techniques that with many layers exploits the data processing in a nonlinear manner to extract features, analyze pattern and classification in a supervised or un-supervised sense. The objectives of deep learning is learning of the features as to their hierarchy, where the higher levels are formed through combined features at lower levels [16]. The latest evolution in deep learning correlated to complex learning task implementations or apply external memory [4, 17, 18] accompanied with reinforcement learning [19].

Reinforcement Learning (RL) is an interactive model different from supervised learning methods. In the latter, the classes' label is determined at training phase, while in RL the discussion on learning is addressed, that is, the agent must learn by itself. All available supervised methods have the similar learning pattern [20, 21]. The learning issue, by merely relying on the available data in reaction of environment is a Reinforcement learning issue. In Reinforcement learning an agent is encountered who interacts with its environment through trial and error, thus, learns to choose the optimal act in achieving its objective [20]. Reinforcement learning is a pattern relevant to learning as to control a system in a sense that its numerical function criterion is maximized as a long-term objective. The difference in Reinforcement learning and supervised learning is only in the partial feedback that the subject receives relevant to its predictions of the environment. Moreover, these predictions might have long term effect, thus, time is a great contributor [22, 23]. In Reinforcement learning the reward and penalize method is applied in the agents' training with no need to determine the performance pattern for the agent [20, 22-24].

This Q learning is one of algorithms mostly applied in reinforcement learning which follows a defined policy to perform different motions in different situations through learning a measure/volume function. One of the strong features of this method is the learning ability of the given subject without having a defined model of the environment [25].

---

[1] http://deeplearning.net/

## 2.2  The Universal Turing Machine

### 2.2.1  The Turing Machine

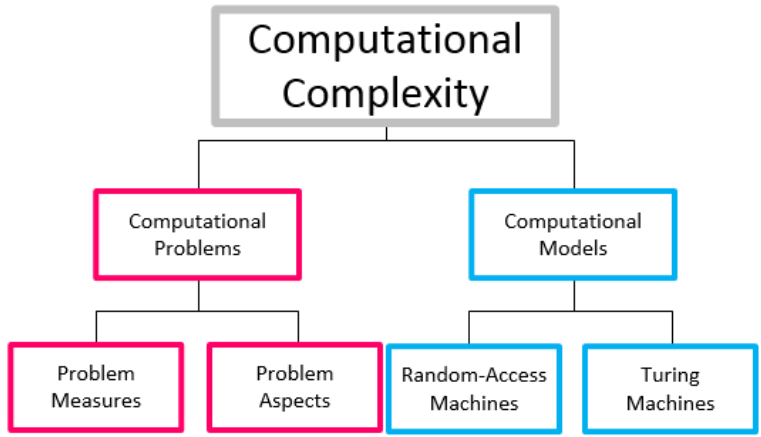

*Fig. 3. The Turing machine mind-map [51-52]*

As observed in Fig. 2 and 3, Turing Machine[26] is a sub-set of computing models. This Turing model is capable to respond to the following two questions: 1) Is there any machine able to determine whether to stop or continue its function through its own memory tapes? 2) Is there any machine able to determine that another machine can ever prints symbol on its tape [27].

In general, the Turing machine is a sample of CPU that controls all of the operations that can be done with the computer on the data and stores the data continuously using a memory. This machine, in specific, is able to work with reliable alphabetic string [27-30]. Turing machine in mathematical sense is a machine that operates on one tape which contains symbols that this machine can read and write and apply both in a simultaneous manner. This action is defined in its complete sense through a series of simple and limited directives. This is a machine equipped with a number of finite states that on any passage prints an element on the tape [31], Fig. 4.

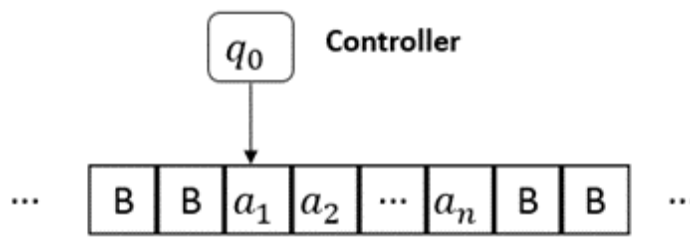

*Fig. 4. A sample of Turing tape*

An unofficial definition of Turing machine: A machine that at the first glance can be considered as a language receptor as well as a computing machine. It consists of a state controller and a memory tape of infinite length and a pointer that points out the portion of the memory that is being read. The controller at every stage, based on the current state and the data it receives from the pointer, performs an act including memory updating and right-left motion on it. At the beginning the machine is set at state $q_0$ and the tape containing the entry string of w=$a_1$, … $a_n$ is surrounded with infinite protracted empty symbols on both its sides [32, 33].

An official definition of Turing machine: A machine of seven state (7-state) ordered in a M= (Q, Σ, Γ, $\delta$, $q_0$, B, F) in which, Q is the total states of the machine, Σ is the entering alphabetic string, Γ is the memory alphabetic tape (where $\Sigma \subseteq \Gamma$), $\delta$ is the function (minor) state transfer where, $\delta : Q \times \Gamma \longrightarrow Q \times \Gamma \times \{R, L\}$, is the initial state of $q_0$, B is a blank symbol, $B \in \Gamma - \Sigma$ and F is the set of final states [32, 33]. There exist two methods to accept the language string for Turing machine: acceptance with achieving final state and acceptance by stoppage.

The question of finding the smallest possible universal Turing Machine with respect to the number of states and symbols is addressed by [34]. The researchers in [35-38] competed on finding the Turing Machine. The initial attempts of [38, 39] led to developing a Turing Machine that simulated the Turing machine in an effective and applicable manner (in a multinomial time) [40]. A small Turing Machine of 4-symbol and 7-state was developed by Minsky in [39] which worked by the simulation of 2-tag systems. This machine was advanced by Rogozhin et al. from its existing state to small machines with several paired symbol-state [41]. The same researcher, with the cooperation of authors in [41-43] improved or made the machines smaller. The machines are shown in Figure 2 and listed in Table 1, entirely.

### 2.2.2 Super Turing Machine and Human Turing Machine

Neural computation is a research base in realizing human brain on an information system. This system, by applying different sensors reads the inputs in a constant manner and codifies the data into different biological variables. Data storage is of the different types of memory (like STMT, LTM, and association memory), that where these operations are named "computation" and the outputs to different channels like control engine of order, decision making, and thoughts and feelings [44]. In the recurrent neural networks, logical weighting leads to continuous improvement in computations and can make them equivalent (same-value) of the Turing machine [3]. On one hand every function can be computed through Turing machine and yet on the other every Turing machine can be simulated in a linear time through a number of recurrent neural networks [45].

The question regarding brain functionality which is analog, parallel and vulnerable in making errors in computing multi-stage computation away from biological noises is addressed in [46]. In [11] a hybrid neural machine is introduced to simulate a person's neural processing system, where every stage includes condensed and parallel computations. According to [30] the Turing machine is introduced as both a similar version of 'a pattern applied by human in running computation on real number' and 'a human with necessary limited memory', an inspiring source for Turing in realizing his own aware cognition. Many researchers do not consider Turing neural sciences as a model for human brain, whereas in [11] the opposite holds true. Which their goal reducing the gap between psychology theories of mental and neurophysiological research by providing an experimental hypothesis as a serial multi-step computation that can be done by parallel circuits of the brain.

In many cognitive architectures, the following three inherited feature through frameworks and share them are determined: 1) coexistence of a big parallel data distribution system, a production system, 2) a selected serial of the products (its on measures), and 3) volume for selected products to determine the system state from sensory to memory, initiating a new repetitious cycle [11]. They have capitalized on this ideas and recommendations to implement an acceptable neural implementation.

## 2.3 Neural networks

Neural network is a type of modeling from real neuron system by means of patterning based on human mind with many applications in solving different problems in different scientific disciplines and the fields of image processing, theme recognition, controlling systems, robotic etc. [1, 47, 48]. Application span of these networks is vast and includes classification, interpolation approximation, detection, etc. with the advantage of capabilities next to easy application. The fundamental base in neural network computations is the human brain features' modeling in a sense that the inspiration thereof would lead to the attempts in formulating the relationship between input and output variables based on the observable data. The general pattern of neural network consist of: 1) reviewing a process in elements named neurons, 2) data interaction through their interconnection, 3) One of these connections has a weight which is multiplied into the data transferred from a neuron to another and this weights is the necessary data for solving problems, and 4) each one of the neurons imposes a activation function to its input to compute the output [49-51]. The general schematic of the neural network categorization is drawn in Fig. 5.

Two main sections in neural network consist of: 1) Training the network: a process where the connecting weights are optimized in a continuous manner in order to minimize the difference between observable and computing values in supervised (applying input and output refers) and unsupervised (no need for objective vector for output) manners, and 2) Validation performance : that is, to assess the network learning and functionality (network capability level regarding its reaction against training input and new inputs).

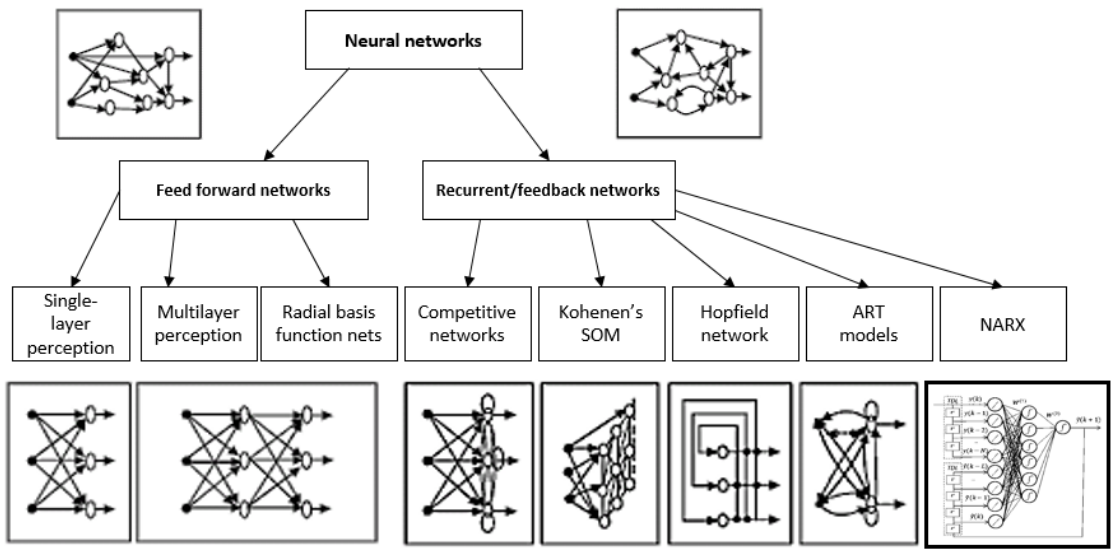

*Fig. 5. Neural Networks*

### 2.3.1 Feed forward Neural Network

Any artificial neural network unable to make direct connection among units by ring or circuit is referred to feed forward Neural Network. This network is the simplest type of systemized artificial neural networks [52]. As observed in Fig. (5), the data moves forward in one direction from input nodes, through hidden nodes (if available) towards the output nodes. There exists no ring or cycle in this network [52, 53].

The neural networks are usually described according to their layers, where each layer in a parallel manner includes input, hidden and output nodes. The most simple feedforward network includes two input nodes and one output node, applicable in modeling logical gates [53]. The multi-level feedforward networks are able to approximate any measurable function at any desired degree of accuracy and precession [54].

### 2.3.2  The hybrid LSTM machine

In fact the Long Short Term Memory (LSTM) is a special type of (RNN) architecture designed to model long-range dependencies with a higher accuracy in relation to regular (RNN) [55, 56] introduced by [55]. It can be claimed that LSTM is the most successful type of architecture regarding RNN for many tasks with data sequence.

LSTM and regular RNN are usually applied in different task sequence prediction and tagging sequences [55]. Learning data storage through Back propagation Recurrent is time consuming due to insufficient errors in back propagation and deterioration feedback. LSTM is able to learn in a shorter time. In comparison with other neural networks, LSTM is able to act faster with higher accuracy and solve the complex and artificial tasks regarding time delay, something not happened before through none of the recurrent network algorithms [57]. Many attempts are made to improve LSTM. One recommendation is to apply working memory in LSTM, which would allow the memory cells in different blocks to correspond and the internal computations to be made in one layer of the memory [4, 58].The standard and improved LSTMs are shown in Fig. (6).

Of course both the LSTM neuron and Turing machines networks face some restrictions in performing this tasks, (e.g. Turing machine is only a computation model for which facing restrictions in physical systems is inevitable [59].

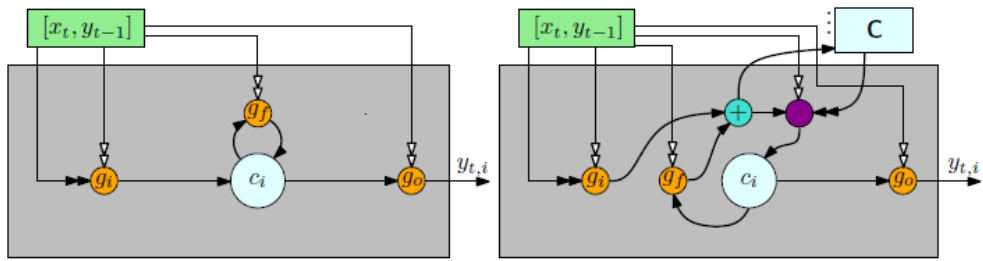

*Fig. 6. Standard LSTM on the left and the recommended one as LSTWM on the right [58]*

### 2.3.3  The NARX neural network

The special recurrent neural network [61] and NARX networks are the networks that model Turing machine and able to simulate any Turing machine by applying the sigmoid activation function [62, 63]. Two schemes of the NARX neural network's architecture are shown Fig. 7. Another type of these networks is the LSTM. The evolution of neural back propagation network leads to production of advanced Turing Machine (super-Turing) with the capability of simulating any definite interactive systems.

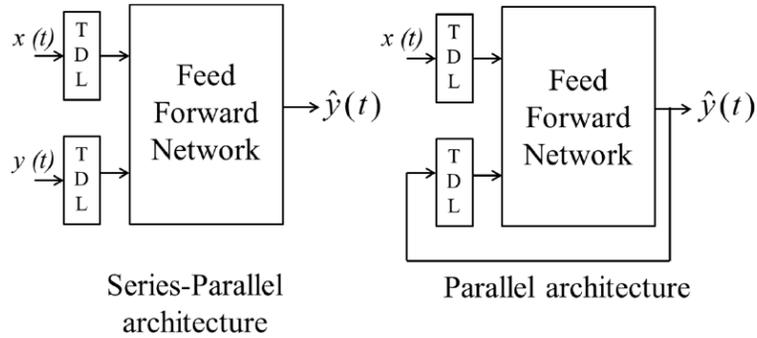

## 3    Turing and Neural Networks

In reality, an infinite network in terms of size can perform the same computation equivalent to a Turing machine. It is expected that to be a good equivalence between neural networks and finite state machines [64, 65]; although, there is no formal proof for this matter. The universal machines are referred to as Turing machines and articles on intelligent machines entails how to make an intelligent machine [66, 67]. There is still problems in lack of memory in neural networks, difficulties in learning of Turing machines etc. All the accomplishments so far in this respect are classified as follows:

### 3.1  Neural Turing Machine (NTM)

One of the main cognitive elements of human is the ability to store and apply information as it is deduced [68-70]. Regardless of all advances made in machine learning [15, 71], the expansion manner of the intelligent factors with similar long-term memory is still ambiguous [68]. It may be claimed that Jordan Pollak [61] is the first to address application of the neural network idea as to its capability in solving problems. He demonstrates that a specific recurrent network model which is named 'Neural Machine' for 'Neural Turing' is universal in this model, all neurons are updated by applying their previous operational volumes in a simultaneous manner [64]. In 2014 Graves, et al. introduced a model named Neural Turing Machine (NTM). This model together with the Memory Networks reveal that addition of one external memory to the recurrent Memory Networks can highly improve functionality [72].

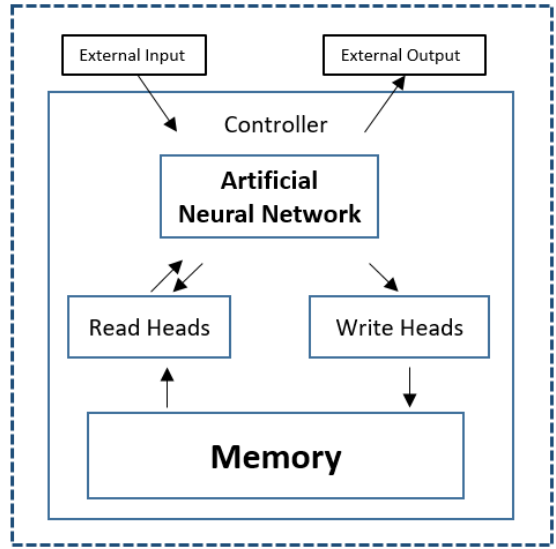

The NTM is a model in deep learning and its success is due to neural network capable of providing correct operation and the ones with learning abilities. These networks have better potential in being promoted provided gaining depth during their main training and having low number of parameters (Simple scheme is shown Fig. 8). The first model of this kind is NTM [73].This is a universal model trainable through a backpropagation algorithm by applying input-output examples [74]. The NTM consists of RNNs with an addressable external memory [4], which improves its capabilities in performing pre-complicated algorithmic tasks like sorting. Most authors are inspired by the cognitive scene through which claim that human is equipped with a central executive system interacting with one memory buffer [6]. In comparison with a Turing Machine, NTM in a program format leads the reading and writing heads in a tape form when direct connection with the external memory is of concern [70]. This ability can be involved in different processes, in fact this hybrid system unlike the Turing Machine or Von Neumann due to this ability in end-to-end diagnosis can be trained through Gradient Decent method [6].

The two main sections of NTM consists of one controller and on memory matrix, Fig. 9. The controller can be of a Feed forward or Recurrent Neural Network, which returns the output to the external world by training the network and receives inputs and reads from memory. The memory is formed by a large matrix of N memory space, each with *m* dimension vectors. Existence of reading and writing facilities allows easy interaction among the control and memory matrix sections [70], Fig. 9.

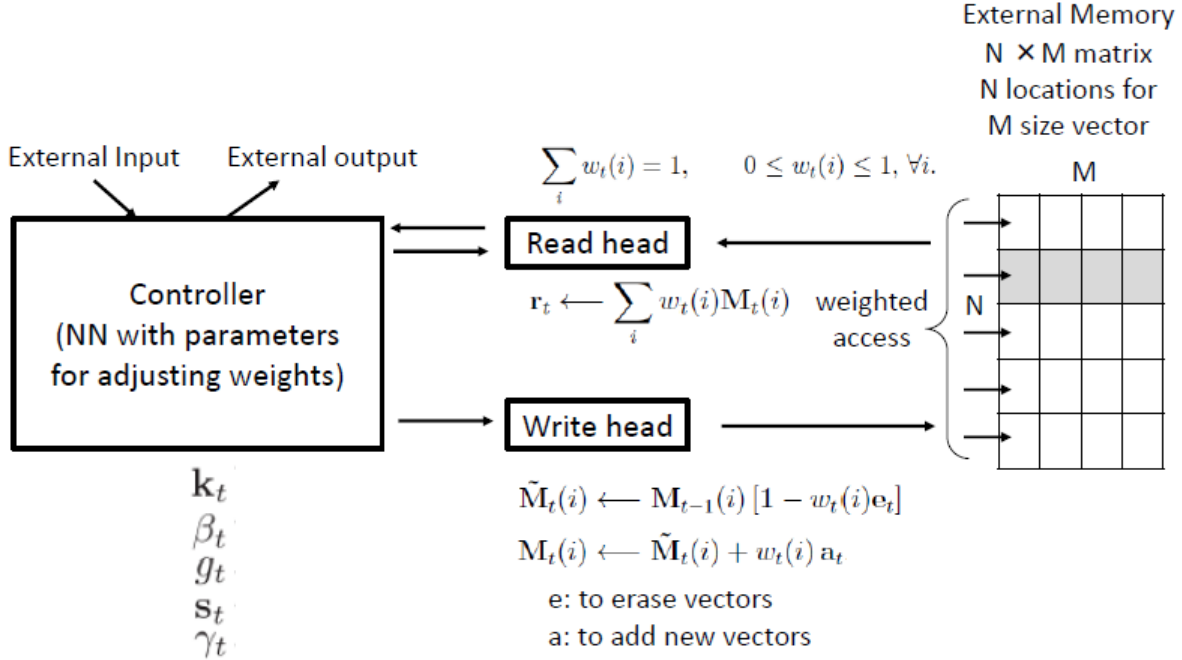

Graves et al [4] selects five algorithm tasks to examine NTM efficiency, here algorithmic means that for each task, the final output receives for the input, be computable through a simple program and be easy in being implemented through any of the common languages. The initial results in this article reveal that Neural Turing machines are able to have a deduction on algorithms like Copying, Sorting and Associative recall with inputs and outputs samples. For example, to copy task, the input of which is a sequential binary vector with fixed length and a limit number of symbols and the objective of the output is to provide a copy of the protracted input. As for sorting which takes place based on priority sort where the input includes a sequence input from the binary vectors together with a priority numeric value determined for each factor and the lengthy inputs in sequence sort of vectors according to their priorities. This test is to measure NTM to see whether it can be trained through supervised learning in order to implement correct and effective algorithmic tasks. The obtained solutions from this method are extended to lengthy inputs compared to the training set, while according to [4, 75], LSTM without external memory is not extendable to lengthy inputs. NTM machines are designed to resolve problems which need rapidly-created variable rules [76]. The computer programs usually apply three fundamental mechanisms: 1) Elementary operations (e.g., arithmetic operations), 2) Logical flow control (branching), 3) External memory. Most modern learning machines do not consider logical flow control and external memory.

The three architectures: 1) LSTM RNN, 2) NTM with a forward facing controller, and 3) NTM with a LSTM controller are assessed in [4]. For each task, both NTM architectures showed better performance than LSTM RNN in both training set and test data generalization as illustrated in Figures 2 to 6. For instance, it is observed that learning in NTM is more rapid than mere LSTM that results in reducing costs; nonetheless, both methods act perfectly.

## 3.2 Reinforcement Learning Neutral Turing

An external memory expands the abilities of NTM and assists solving problems and performing tasks. Other learning machines are unable to perform, that is, a model can be expanded through its relevant extreme connections with the outer world, Fig.10. These external connections are: memory, a database, a search engine or a part of a software like theorem verifier. In this proposed method, the reinforcement learning neutral Turing is applied in training the neural network where interaction of which leads to performing simple algorithmic tasks. Many of these important connections are of expanded data bases and search engines.[74].

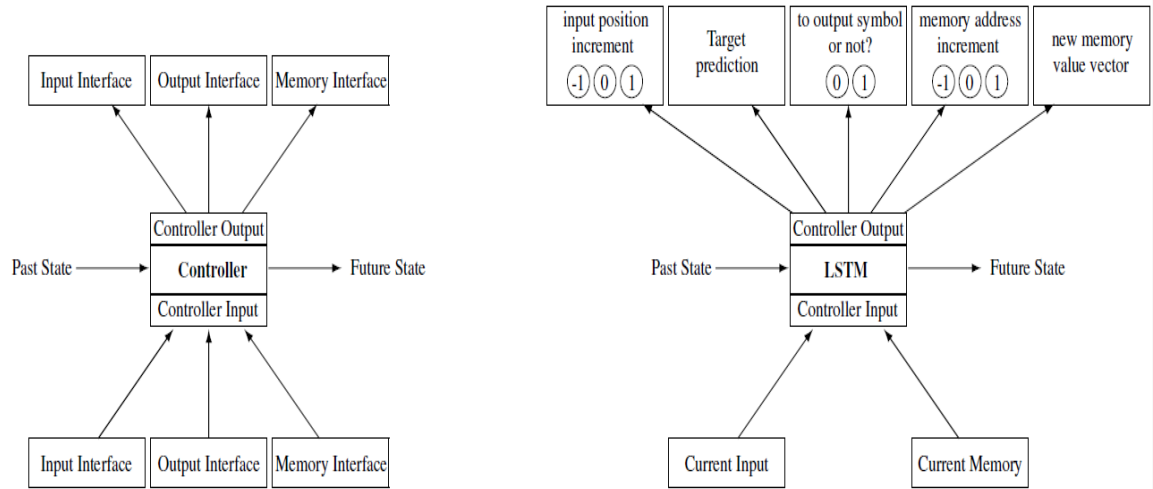

*Fig. 10. Left: a brief scheme of Interface-controller model, Right: a sample of the proposed interface-controller model introduced by [74]*

The reinforcement learning of neural Turing machine model is the first model that is a perfect Turing machines. This does not mean that it can be trained to perform hard tasks. Most of the difficult tasks need lengthy and multi-step connections with an external environment (e.g. the computer games [77], stock exchanges, commercialization systems or the physical world [78]). A model partially observes the environment and influences on its surrounding by actions that can be generally seen as a part of reinforcement learning.

*Table 1 A briefing of whatever the controller reads and produces at any time stage; presenting the learning pattern of each section of the model [74]*

| Interface | | Read | Write | Training Type |
|---|---|---|---|---|
| Input Tape | Head | window of values surrounding the current position | distribution over[-1, 0, 1] | Reinforce |
| Output Tape | Head | ∅ | distribution over[-1, 0, 1] distribution over output vocabulary | Reinforce Backpropagation |
| | Content | ∅ | | |
| Memory Tape | Head | window of memory values surrounding the current address | distribution over[-1, 0, 1] vector of real values to store | Reinforce Backpropagation |
| | Content | | | |
| Miscellaneous | | all actions taken in the previous time step | ∅ | ∅ |

The section of the model that interacts with the errors is named the controller, the only section that is trained. An abstract controller model and the proposed model [74] are shown in Fig. 10. RL-NTM learning through reinforcement algorithm is applied in discrete decision making where it learns to produce outputs using backpropagation. A briefing on what a controller reads and produces at every time stage is tabulated in Table 1. The model required for training is expressed in the train column of this table [74].

### 3.3 Evolving Neural Turing Machines

In this method, instead of adopting training NTM with gradient distant. The evolutionary algorithm is illustrated in Fig. 11 as well. The initial results indicate that this model can make the neuron model more simpler, let it have better generalization and no need to have access the total memory content in every time stage [68, 73].

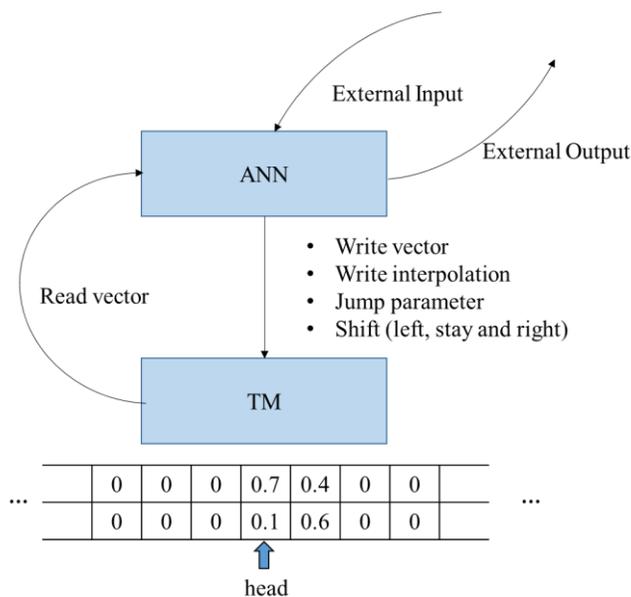

*Fig 11. A scheme of ENTM illustrating the NTM deduction abilities [68]*

### 3.4 Lie Access NTM

The issue of a new type of generalized Neural Turing Machine by the operations of Lie Access (assumed name) group is addressed in [72] and according to tests and comparisons the superiority of LANTM to LSTM at copy, double and reverse tasks are proven.

The primary contributions of this work are the following: 1) Using Lie group actions instead of convolutions to manipulate read and write head; mathematically, this assures that all actions are associative and invertible, and assures the identity transformation, none of which are guaranteed by the original formulation of the NTM (and later ones too). 2) Using manifolds instead of traditional tape-based systems to store memories; this guarantees local smoothness and naturally provides a distance metric. 3) Use of inverse-square and comparison to softmax weighting of memories when reading.

Of course LANTM is tested by the two InvNorm and SoftMax addressing mechanism variables [72].

### 3.5 DNC (Differentiable Neural Computer)

Yet another learning method inspired from NTM model is differentiable neural computer (DNC). In this model, an architecture is used similar to the neural network controller with read/write heads that access to memory matrix. However, the way of communicating with memory is different in that a dynamic memory is used in order to resolve memory limitations. Information are maintained if they are repeated more than a threshold. This method as a regular computer uses memory in order to illustrate and manipulate complex data structures; however, it learns from data similar to neural networks. Applying this method along with supervised learning shows that it is able to respond to synthetic queries on natural language deductions on bAbI dataset [79]. In addition, they have the ability of performing tasks on graphs such as finding the shortest path between location points (such as transportation network), and conjecturing lost links (such as family genealogy) [5].

### 3.6 Dynamic Neural Turing Machine

The advanced NTM model is trainable by defining one addressable memory. In this pattern every memory cell has two separate vectors of address and content allowing a variety of learning strategies to be learned with respect to addressing whether it is a linear or non-linear space. Addressing memory in D-NTM computers is equivalent to computing one of the n-diminution address vector. D-NTM calculates three reading, writing and erasing vectors, specifically for writing the complementary computations of the controller for vector based on Hidden state [10].

### 3.7 The implemented tasks in NTM

All of the tasks that mentioned above will be explain in this section (tabulate in Table 2):

#### 3.7.1 Copying Task

The symbols of input tape are copied on the output strip. In fact, an input sample is received and stored in the memory, and then the same sequence is produced. In reality, this task tests whether NTM is able to store a lengthy sequence of desired information and then remember them. This network is provided by a sequence of random binary vectors together with a separating flag. Storing and accessing to information become troublesome for RNNs and other dynamic architectures after a long time [4, 69, 72-74, 80, 81]. It is notable that, the applied algorithm in the tests has applied both the addressing based on content and space [4, 5].

#### 3.7.2 Repeat copy Task

This is to extend the copying task based on the copied output sequences with determined repetitions, represented by a symbol at the end of the sequence. What is essential here is to realize whether NTM is able to learn a simple nested function. The network receives sequences from binary vectors with random length followed by showing a digital volume of the determined copies.

The networks are trained to re-produce sequences with 8 random binary vectors in which the length of vector and number of iterations are chosen from the range [1-10]. The algorithm input indicates the repetition number, which is normalized with a Zero Mean and varies of one [4, 81].

#### 3.7.3 Associative Recall Task

Here, one item is defined as a protracted binary vectors surrounded by separating symbols on either side. After few items propagate the system, by sharing one random item the network is asked to produce the next item, where each item consists of three 6 bit binary vectors (6 * 3= 18) bits/item and in testing at least 2 and at the most 6 items are applied in each section [4].

### 3.7.4    N-Gram of the Dynamic N-Grams Task

The objective here is to know whether NTM is able to adapt itself to new predicted distribution in a rapid manner and by applying its memory as a re-writable table, hold the number of statistical transforms which would end up in being emulated as a simple N-Gram model.  Here, all 6-Gram distribution on the binary sequence are of concern.

Every 6-Gram distributions can be in the form of a $2^5$=32 number table. Determining the probability that the next bit might be 1, five binary histories are initially presented by 6-Gram random probability in a sense that all 32 distribution probabilities from Beta (½, ½) distribution are drawn in an in dependent manner. This is followed by production of a specific learning sequence by drawing a sequential 200 bit through the available search table which observes a sequence from one bit, this requests the prediction of the next bit. The optimized estimator for this issue can be determined through Bayesian analyses [4, 82].

### 3.7.5    Priority Sort Task

Here, whether NTM is able to test data sorting is tested (an initial essential algorithm). One protraction from random binary vectors is given to every vector as an entry to the network with a degree of scalar priority. This priority is drawn in Zone [-1, 1] in a uniform manner. The objective protraction consists of sorted binary vectors according to priorities. Every input protraction consists of 20 binary vectors with similar priorities with the objective of maximum 16 vectors of priority in the input. For better performance, 8 parallel heads of reading and writing and a feed forward controller are required, indicating the difficulty of sorted vectors reflection trough applying unary vector operations [4].

### 3.7.6    Arithmetic Task

This task is to perform addition or subtraction of two long numbers in input exhibited through digital sequences and each input sequence delimits with the character "]" from other the sequences calculation [83]. For example:

$$-4 - 98307541 = -98307545] - 942 + 1378 = 432] - 11 + 854621 = 854610]\ 125804 + \cdots$$

### 3.7.7    Variable Assignment

This task is designed to assign a variable to test neural network ability for key-value recovery. Here, a hybrid simple language consisting of one protraction of 1-4 value variable assignment sequence through a query formed of Q(variable); therefore, the network should guess the value where the variables' name follows 1-4 character in a random manner [83].

s( ml,a ), s( qc,n ), q(ml)<u>a</u>.s(ksxm, n), s(u,v), s(ikl,c), s(ol,n), q(ikl)<u>c</u>.s(...

### 3.7.8    XML Modeling

The XML language includes involute tags (theme independent) in form. The input is protract consisting of tags with names formed from 1-10 characters in a random manner. The tag name is predictable only when it is closed, $< tag >$, thus, cost predicting in protraction is subject to the tags being closed. Every symbol must be predicted one step prior to being presented. The underlined sections in XML are limited to the maximum number of 4 nested tags' depth to which the coast measured are assigned in order to prevent the intensively long sequences of open tags during production [83].

<u>≤xkw≥</u><u>≤svgquspn≥</u><u>≤oqrwxsln≥</u><u>≤/aqrwxsln≥</u> <u>≤/svgquspn≥</u><u>≤jrcfcacaa≥</u><u>≤/jrcfcacaa≥</u><u>≤/xk</u>…

### 3.7.9 Reverse

The objective is to invert the symbols from input strip to output strip. The specific character '1' is applied to show the protracted end. This task must learn to have rightward orientation until it encounters 'r' and then leftward orientation to copy the symbols in the output [72, 74].

### 3.7.10 Reverse Forward

The performance of this task is similar to that of the Reverse, except that the control is allowed to move forward to the input strip pointer, that is, it must apply to extended memory for a complete solution [74, 81].

### 3.7.11 Duplicate input

The generic input is in $x_1 x_1 x_1 x_2 x_2 x_2 x_3 \ldots x_{c-1} x_c x_c x_c \emptyset$ form, while the output of concern is $x_1 x_2 x_3 \ldots x_{c-1} x_c \emptyset$; accordingly, every entry symbol is repeated three times, allowing the controller to transmit the third symbols of the input [74, 81].

### 3.7.12 Addition

Here, the model learns how to add two (two multi protracted digits) with the same length together. This task consists of 3 actions: 1) Having one adding table for the priority of digits, 2) learning about the moving pattern on input grid and 3) realizing the carry concept [72, 81].

### 3.7.13 Bigram Flip

The source is limited to the length of every protraction. The objective is accomplished by replacing the i[th] symbol with i+1[th] [72, 80].

$$< s > \; a_1 a_2 a_3 a_4 \ldots a_{k-1} a_k \; ||| \; a_2 a_1 a_4 a_3 \ldots a_k a_{k-1} \; </s>$$

### 3.7.14 Double

Here, X is put in a $[0, 10^k]$ zone as an integer together with zero padding in front (on the right side) to form digit K; input: zero padded for digit K, X based on 10 and output: zero padded for digit K+1, X based on 10 [72].

### 3.7.15 bAbI

This episodic task consists of sets of training and Question-answering tests accessible through *http://fb.ai/babi* and is generally named the Feedback bAbI [10, 84].

### 3.7.16 Graph experiment

To investigate the ability of DNCs to exploit their memory for logical planning tasks. in this task, created a block puzzle game inspired by Winograd's SHRDLU30—a classic artificial intelligence demonstration of an environment with movable objects and a rule-based agent that executed user instructions. for which the networks were trained with supervised learning, we applied a form of reinforcement learning in which a sequence of instructions describing a goal is coupled to a reward function that evaluates whether the goal is satisfied—a set-up that resembles an animal training protocol with a symbolic task cue.

| Tasks \ Methods | NTM [4, 72, 83] | RL-NTM [85] | ENTM [68] | LANTM [72] | ALSTM [83] | D-NTM [10] | LSTM [4, 72, 83] | DNC [86] |
|---|---|---|---|---|---|---|---|---|
| Addition | ✓ | – | – | ✓ | – | – | ✓ | – |
| Arithmetic | ✓ | – | – | – | ✓ | – | ✓ | – |
| Assignment | ✓ | – | – | – | ✓ | – | ✓ | – |
| Associative | ✓ | – | – | – | – | – | ✓ | – |
| bAbI | – | – | – | – | – | ✓ | – | ✓ |
| Bigram Flip | – | – | – | ✓ | – | – | ✓ | – |
| Copy | ✓ | ✓ | ✓ | ✓ | ✓ | – | ✓ | ✓ |
| Double | – | – | – | ✓ | – | – | ✓ | – |
| Duplicate Input | – | ✓ | – | – | – | – | – | – |
| Forward Reverse | – | ✓ | – | – | – | – | – | – |
| Graph | – | – | – | – | – | – | – | ✓ |
| N-Gram | – | – | – | – | – | – | ✓ | – |
| Priority Sort | ✓ | – | – | – | – | – | ✓ | – |
| Repeat Copy | ✓ | ✓ | – | – | – | – | ✓ | – |
| Reverse | – | ✓ | – | ✓ | – | – | ✓ | – |
| T-Maze | – | – | ✓ | – | – | – | – | – |
| XML | ✓ | – | – | – | ✓ | – | ✓ | – |

## 4    DISCUSSION

### 4.1  Comparison of implemented tasks in different controllers

#### 4.1.1    Comparison of implemented tasks in NTM and LSTM

The results of the implemented tasks obtained through NTM and LSTM consist of Copy, Repeat Copy, N-Grams, Priority Sort and Associative Recall, illustrated in Figs. (12-17) where the outperformance of NTM is relation to LSTM is expressed as to learning speed and output production (i.e. in Fig. (12), the task (copy) is performed faster and better, thus with low error in relation to LSTM [4]. The adjustments regarding the controller size, head count, memory size, learning rate and parameters count applied in NTM and LSTM are tabulated in Table 3-5 for test purposes.

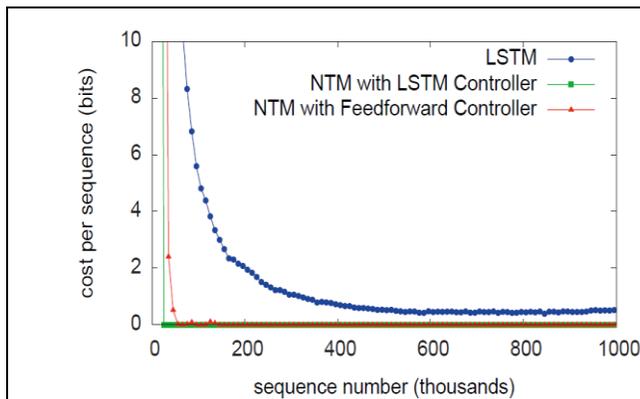

*Fig. 12. learning curve, Copy task [6]*

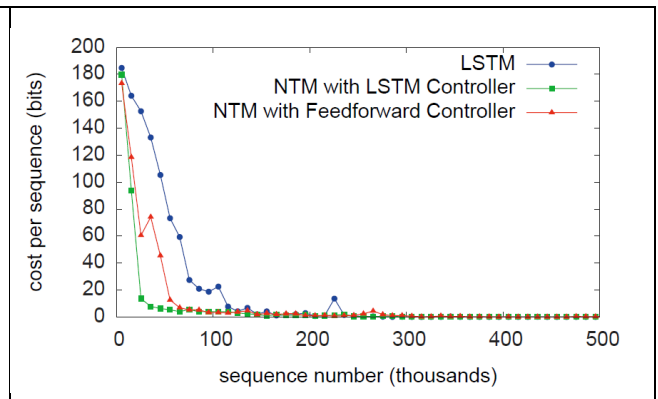

*Fig. 13. Learning curve, Repeat copy [6]*

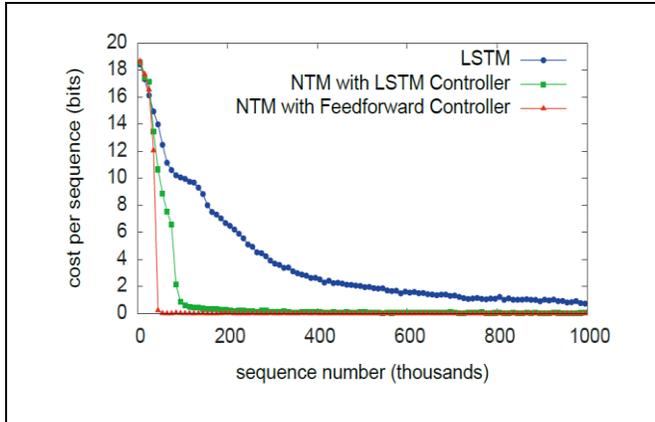

*Fig. 7. Learning curve of LSTM, Associative Recall[6]*

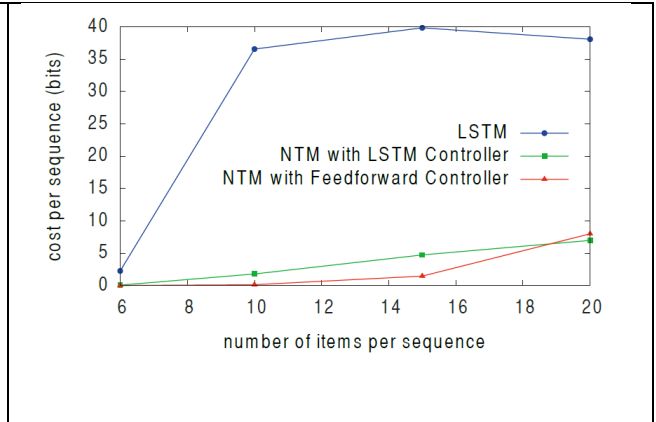

*Fig. 8. The generalized function on Associative Recall for lengthier protractions [6]*

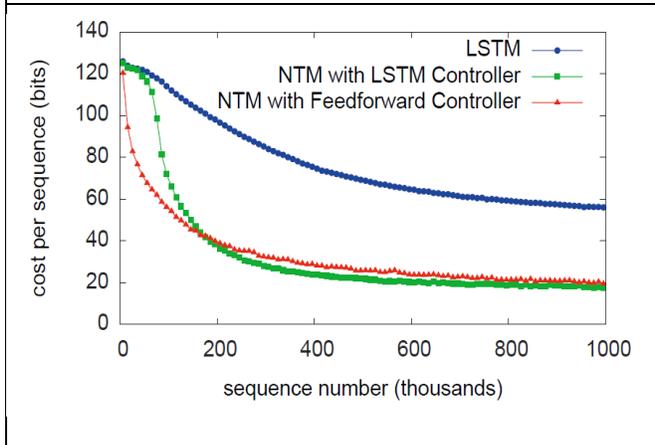

*Fig. 9. Learning curves, Priority sorting [6]*

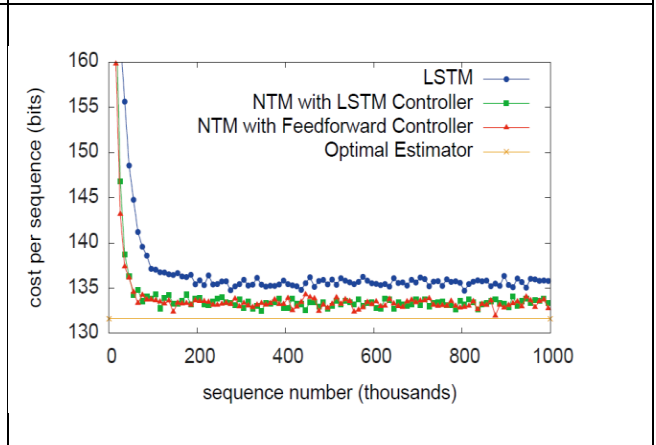

*Fig. 107. Associative Recall, Dynamic N-Grams [6]*

*Table 3 Adjustment of experimental NTM with progressive controller [6]*

| Task | #Heads | Controller Size | Memory Size | Learning Rate | #Parameters |
|------|--------|-----------------|-------------|---------------|-------------|
| Copy | 1 | 100 | 128×20 | $10^{-4}$ | 17,162 |
| Repeat Copy | 1 | 100 | 128×20 | $10^{-4}$ | 16,712 |
| Associative | 4 | 256 | 128×20 | $10^{-4}$ | 146,845 |
| N-Grams | 1 | 100 | 128×20 | $3 \times 10^{-5}$ | 14,656 |
| Priority Sort | 8 | 512 | 128×20 | $3 \times 10^{-5}$ | 508,305 |

*Table 4 Adjustment of experimental LSTM controller [6]*

| Task | #Heads | Controller Size | Memory Size | Learning Rate | #Parameters |
|------|--------|-----------------|-------------|---------------|-------------|
| Copy | 1 | 100 | 128×20 | $10^{-4}$ | 67,561 |
| Repeat Copy | 1 | 100 | 128×20 | $10^{-4}$ | 66,111 |
| Associative | 1 | 100 | 128×20 | $10^{-4}$ | 70,330 |
| N-Grams | 1 | 100 | 128×20 | $3 \times 10^{-5}$ | 61,749 |
| Priority Sort | 5 | 2×100 | 128×20 | $3 \times 10^{-5}$ | 269,038 |

*Table 5 Adjustment of experimental LSTM network [6]*

| Task | Network Size | Learning Rate | #Parameters |
|---|---|---|---|
| Copy | 3 ×256 | $3 \times 10^{-5}$ | 1,352,969 |
| Repeat Copy | 3 ×512 | $3 \times 10^{-5}$ | 5,312,007 |
| Associative | 3 ×256 | $10^{-4}$ | 1,344,518 |
| N-Grams | 3 ×128 | $10^{-4}$ | 331,905 |
| Priority Sort | 3 ×128 | $3 \times 10^{-5}$ | 384,424 |

## 4.1.2    Comparing both the LANTM and LSTM for different tasks

Table 6. The results of test: 1X is the input count equal to training input, 2X is the input count twice the training inputs [72].

*Table 6 The LANTM-InvNorm generalized results on more input count [72]*

| Task | model | 1x coarse | 1x fine | 2x coarse | 2x fine |
|---|---|---|---|---|---|
| Copy | LANTM-InvNorm | 100% | 100% | 100% | 100% |
| | LANTM-SoftMax | 100% | 100% | 99% | 100% |
| | LSTM | 58% | 97% | 0% | 52% |
| Reverse | LANTM-InvNorm | 100% | 100% | 100% | 100% |
| | LANTM-SoftMax | 1% | 12% | 0% | 4% |
| | LSTM | 65% | 95% | 0% | 44% |
| BigramFlip | LANTM-InvNorm | 100% | 100% | 99% | 100% |
| | LANTM-SoftMax | 12% | 40% | 0% | 10% |
| | LSTM | 98% | 100% | 4% | 58% |
| Double | LANTM-InvNorm | 100% | 100% | 100% | 100% |
| | LANTM-SoftMax | 100% | 100% | 100% | 100% |
| | LSTM | 98% | 100% | 2% | 60% |
| Addition | LANTM-InvNorm | 100% | 100% | 99% | 100% |
| | LANTM-SoftMax | 17% | 61% | 0% | 29% |
| | LSTM | 97% | 100% | 6% | 64% |

With respect to the content of Table 6. It is observed that LANTM-InvNorm outperforms LSTM regarding all tasks and their generalizations for twice size of learning input data. LANTM-SoftMax outperforms LSTM on Copy and Double tasks while it fails on the rest of the tasks.

The satisfactory result on LANTM-InvNorm provide the means to test the given data up to 8x and the findings are tabulated in Table 7.

*Table 3 The results of generalized LANTM-InvNorm based on increase input [72]*

| Task | 4x coarse | 4x fine | 5x coarse | 5x fine | 8x coarse | 8x fine |
|---|---|---|---|---|---|---|
| Copy | 100% | 100% | 91% | 100% | - | - |
| Reverse | 91% | 98% | 12% | 65% | - | - |

| BigramFlip | 96% | 100% | - | - | 12% | 96% |
| Double | 86% | 99% | - | - | 21% | 90% |
| Addition | 2% | 95% | - | - | - | - |

### 4.1.3   The three NTM, ALSTM and LSTM methods compared

The comparison is made in [83] and for the three XML, Arithmetic and Assignment tasks the following 3 Figs(18,19,20) are yield:

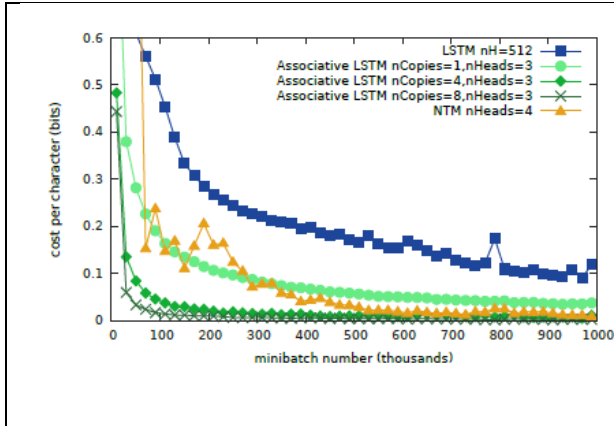

*Fig. 18. Training cost, XML [83]*

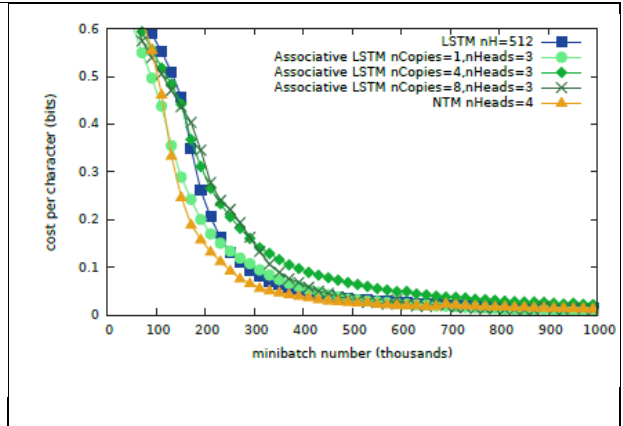

*Fig. 19. Training cost, Arithmetic [83]*

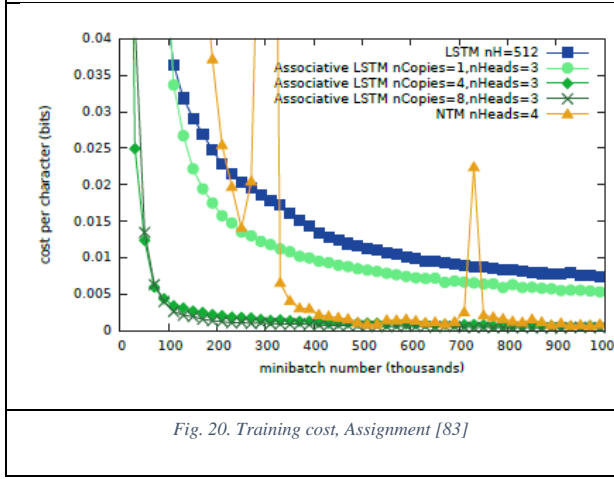

*Fig. 20. Training cost, Assignment [83]*

Both the (ALSTM and LSTM) networks are compared with Neural Turing Machine in Table 8.

*Table 4 Networks compared against NTM [83]*

| Network | Memory Size | Relative speed | #parameters |
|---|---|---|---|
| LSTM nH=512 | $N_h$=512 | 1 | 1,256,691 |
| Associative LSTM nHeads=3 | $N_h$=128(=64 complex numbers) | 0.66($N_{copies}$=1)<br>0.56($N_{copies}$=4)<br>0.46($N_{copies}$=8) | 775,731 |
| NTM nHeads=4 | $N_h$=384, memorySize = 128 x 20 | 0.66 | 1,097,340 |

### 4.1.4 Comparison of Copy and Addition tasks in some methods against NTM

The comparison yield from some of the studies conducted on external memory and algorithmic learning tasks between Copy and Addition is tabulated in Table 9, [72]. The columns here include volumes in the form of A[;B], where A is the input volume with a 99% learning accuracy for both fine and coarse states and be if applied could be the same as A [72].

*Table 9 The comparison of some methods against NTM [72]*

|  | Copy | Vocab | Addition |
|---|---|---|---|
| LSTM | 35 | 10 | 9 |
| NTM | 20;20 | 256 | - |
| LANTM-InvNorm | 64;128 | 128 | 16;32 |

The adjustments made in Tables 9, 10 and 11 reveal that LANTM-InvNorm yield between results in relation to other methods with respect to time performance. More capabilities are observed in conducting different tasks with bigger data in less time with high accuracy when deep learning is applied. Due to its high potential, the NTM is adopted and recommended in may PhD dissertations and articles (e.g. the methods proposed in [84] regarding NTM and NRAM are to assess the operational differences between neural networks and computer programs).

The authors in [87], inspired by the NTM model introduced by [4] and the network memory in [17], by sharing an external memory introduced the two Global Shared Memory and Local Global Hybrid Memory to solve multi-task learning problems and their experimental results indicate that in a shared learning context can improve the performance of each task in relation independent manner.

## 4.2 Method comparison, structural perspective

In general the concept of NTM is very strong but restricted by weak training algorithms [88]. Due to its memory, it is able to outperform many recurrent neural networks. The structural adjustments regarding the methods discussed in this article are tabulated in Table 10, where all have external memory and the neural network is applied as the controller. For training them reinforcement, ultra innovative, Backpropagation and RMSProp algorithms are applied which have yield variety of methods, some of which outperform the main NTM method according to the findings in Sec. 4-1. The comparisons are made against one of the best recurrent neural networks, the (LSTM) and all have outperformed (LSTM). The structural adjustments of LSTM expressed in Table 11. Of course by introducing the ALSTM method in Sec. 4-1-3 it is observed that it is able to outperform NTM on the assignment task to a certain degree, thus, it is expected that if the training method or the controller of NTM is improved, this draw back would be removed.

| Method | External memory | Training with | Controller | Tasks | Controller Size | Memory Size | Learning Rates | Parameters |
|---|---|---|---|---|---|---|---|---|
| NTM [4] | ✓ | Backpropagation | LSTM | Copy | 100 | 128×20 | $10^{-4}$ | 17,162 |
| | | | | Repeat Copy | 100 | | $10^{-4}$ | 16,712 |
| | | | | Associative | 256 | | $10^{-4}$ | 146,845 |
| | | | | N-Grams | 100 | | $3 \times 10^{-5}$ | 14,656 |
| | | | | Priority Sort | 512 | | $3 \times 10^{-5}$ | 508,305 |
| | | | feed forward neural network | Copy | 100 | 128×20 | $10^{-4}$ | 67,561 |
| | | | | Repeat Copy | 100 | | $10^{-4}$ | 66,111 |
| | | | | Associative | 100 | | $10^{-4}$ | 70,330 |
| | | | | N-Grams | 100 | | $3 \times 10^{-5}$ | 61,749 |
| | | | | Priority Sort | 2×100 | | $3 \times 10^{-5}$ | 269,038 |
| ENTM [89] | ✓ | Evolutionary algorithm | ANN | Copy Continuous T-Maz | _____ | _____ | _____ | _____ |
| RL-NTM [85] | ✓ | Backpropagation Reinforcement Learning | LSTM | Copy Duplicated Input Reverse Repeat Copy Forward Reverse | _____ | _____ | Fixed 0.05 & momentum is 0.9 | _____ |
| D-NTM [10] | ✓ | backpropagation | GRU or feedforward or recurrent | Facebook bAbI | _____ | 128 | _____ | _____ |
| LANTM InvNorm [90] | ✓ | RMSProp | LSTM | Copy | 50 | Vocab* | 0.02 | 26,105 |
| | | | | | | 128 | | |
| | | | | Reverse | 50 | 128 | 0.02 | 26,105 |
| | | | | Bigram Flip | 100 | 128 | 0.02 | 70,155 |
| | | | | Addition | 50 | 14 | 0.01 | 20,291 |
| | | | | Double | 50 | 14 | 0.02 | 18,695 |
| LANTM SoftMax [90] | ✓ | RMSProp | LSTM | Copy | 50 | 128 | 0.02 | 26,156 |
| | | | | Reverse | 50 | 128 | 0.02 | 26,156 |
| | | | | Bigram Flip | 100 | 128 | 0.02 | 72,123 |
| | | | | Addition | 50 | 14 | 0.01 | 20,291 |
| | | | | Double | 50 | 14 | 0.02 | 20,291 |

| Method | Tasks | Controller Size | Memory Size | Learning Rates | Parameters |
|---|---|---|---|---|---|
| LSTM [4, 83, 90] | Copy | Network Size | _____ | $3 \times 10^{-5}$ | 1,352,969 |
| | | 3×256 | | | |
| | Repeat Copy | 3×512 | | $3 \times 10^{-5}$ | 5,312,007 |
| | Associative | 3×256 | | $10^{-4}$ | 1,344,518 |
| | N-Grams | 3×128 | | $10^{-4}$ | 331,905 |
| | Priority Sort | 3×128 | Vocab* | $3 \times 10^{-5}$ | 384,424 |
| | Reverse | 4×256 | 128 | $2\times10^{-4}$ | 1,918,222 |
| | Bigram Flip | 4×256 | 128 | $2\times10^{-4}$ | 1,918,222 |
| | Addition | 4×256 | 14 | $2\times10^{-4}$ | 1,918,222 |
| | Double | 4×256 | 14 | $2\times10^{-4}$ | 1,918,222 |
| ALSTM | _____ | _____ | 128 | ____ | 775,731 |

# 5    Recommendations and future works

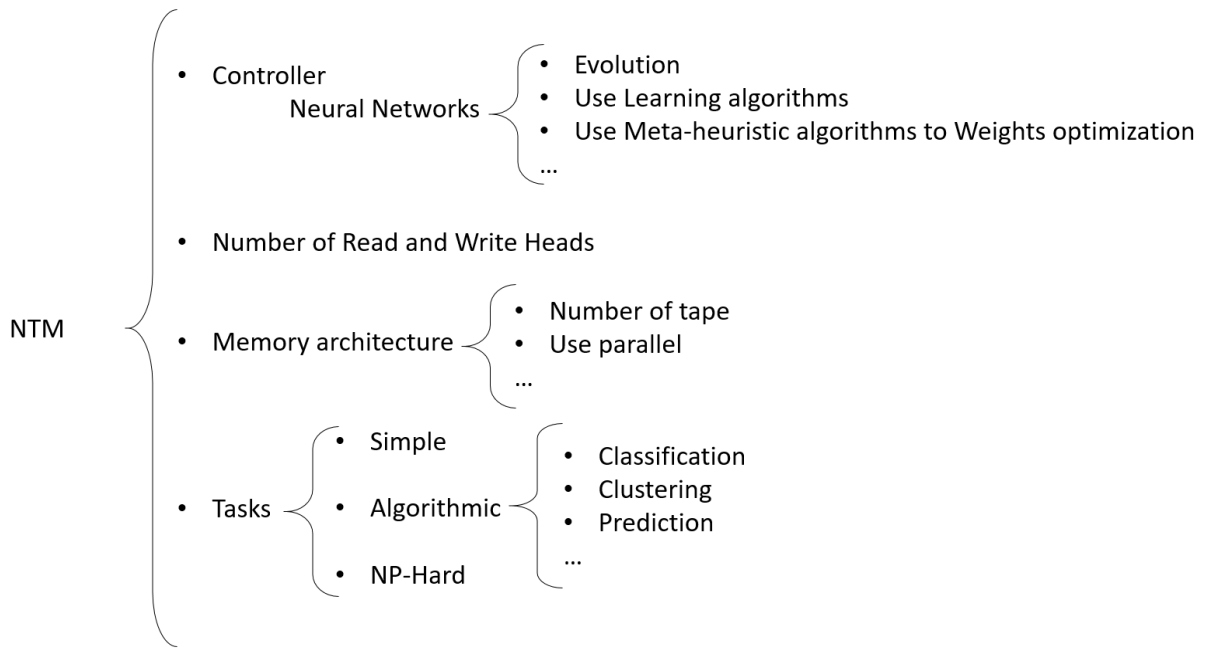

*Fig. 21. Categorization of NTM Methods*

The studies in the field of deep learning are ongoing and due to neural networks' high potential, attempts are made to improve variety of learning methods by adding external memory to them. Can categorize the NTM methods like Fig. 21. based on the parts of NTM and there exists a verity of fields in which NTM is presumably well fitted some of which are as follows. 1) Studying of the relation among different tasks with suitable structure of NTM is worth studying. 2) Studying how to run several NTM in parallel to emulate the behavior of a human brain that do several jobs simultaneously. 3) How to share information among several NTM instances is of crucial importance. 4) It is predicted that through some structural modifications the performance of NTM would improve in future. 5) Improvements in NTM controller or applying different learning methods for it (use another Neural Networks like sections 3-2 and 3-3 which use Reinforcement Learning and Evolving Algorithm for training the NTM). 6) using another structure for memory and memory tapes (like section 3-5, DNTM) and promoting the Turing Machine. 7) Implementation of NTM based on High Performance Computing (HPC) and High Transactional Computing (HTC). 8) In addition to its high capacity in performing complicated tasks, it is applied in solving problems and performing tasks in the realm of: A) Machine learning tasks such as various clustering, classification, association and prediction tasks. B) Tasks in prominent areas such as cryptography, information hiding, etc. C) Tasks in image processing, computer vision, image caption, etc. D) Tasks in voice detection and natural language processing, etc.

## 6    Author biography

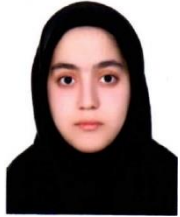 **Soroor Malekmohammadi-Faradounbeh** received her MSc. degree (Artificial Intelligence) in 2018 from Islamic Azad University, Najafabad Branch, and received her B.E. degree (Software Engineering) in 2013 from Islamic Azad University, Mobarakeh Branch. Her research interests are intelligent computing, Machine Learning and Neural Networks, Autonomic Computing, and Bio-inspired Computing.

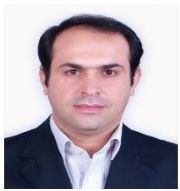 **Faramarz Safi-Esfahani** received his Ph.D. in Intelligent Computing from University of Putra Malaysia in 2011. He is currently on faculty of Computer Engineering, Islamic Azad University, Najafabad Branch, Iran. His research interests are intelligent computing, Big Data and Cloud Computing, Autonomic Computing, and Bio-inspired Computing.